\documentclass[twocolumn]{IEEEtran}
\usepackage{lscape}
\usepackage{booktabs}
\usepackage{authblk}
\usepackage{url}
\usepackage{silence}
\usepackage[font=scriptsize]{caption}

\usepackage[pdftex]{graphicx}
\usepackage{epstopdf}
\usepackage{pifont}
\usepackage[table,xcdraw]{xcolor}
\usepackage{array}
\usepackage{verse}
\usepackage{longtable}
\usepackage{supertabular}
\usepackage{cite}
\usepackage{color}
\usepackage{amsmath}
\usepackage{lettrine}
\usepackage{hyperref}
\usepackage{multirow}
\usepackage{multicol}
\usepackage{dirtytalk}
\usepackage{color}
\usepackage{algorithm} 
\usepackage{algorithmic}  
\usepackage[algo2e]{algorithm2e}
\usepackage[TABBOTCAP]{subfigure}

\begin{document}

\title{Collaborative Federated Learning For Healthcare: Multi-Modal COVID-19 Diagnosis at the Edge}

\author{Adnan Qayyum$^1$\thanks{Email: adnan.qayyum@itu.edu.pk}, Kashif Ahmad$^2$, Muhammad Ahtazaz Ahsan$^1$, Ala Al-Fuqaha$^2$, and Junaid Qadir$^1$ \\ \vspace{2mm}
$^1$ Information Technology University (ITU), Punjab, Lahore, Pakistan \\ 
$^2$ Information and Computing Technologies (ICT) Division, College of Science and Engineering (CSE), Hamad Bin Khalifa University, Doha, Qatar}
\maketitle

\begin{abstract}
\label{abstract}
Despite significant improvements over the last few years, cloud-based healthcare applications continue to suffer from poor adoption due to their limitations in meeting stringent security, privacy, and quality of service requirements (such as low latency). The edge computing trend, along with techniques for distributed machine learning such as federated learning, have gained popularity as a viable solution in such settings. In this paper, we leverage the capabilities of edge computing in medicine by analyzing and evaluating the potential of intelligent processing of clinical visual data at the edge allowing the remote healthcare centers, lacking advanced diagnostic facilities, to benefit from the multi-modal data securely. To this aim, we utilize the emerging concept of clustered federated learning (CFL) for an automatic diagnosis of COVID-19. Such an automated system can help reduce the burden on healthcare systems across the world that has been under a lot of stress since the COVID-19 pandemic emerged in late 2019. We evaluate the performance of the proposed framework under different experimental setups on two benchmark datasets. Promising results are obtained on both datasets resulting in comparable results against the central baseline where the specialized models (i.e., each on a specific type of COVID-19 imagery) are trained with central data, and improvements of 16\% and 11\% in overall F1-Scores have been achieved over the multi-modal model trained in the conventional Federated Learning setup on X-ray and Ultrasound datasets, respectively. We also discuss in detail the associated challenges, technologies, tools, and techniques available for deploying ML at the edge in such privacy and delay-sensitive applications.    

\end{abstract}

\section{Introduction}
The main motivation for using data storage and computing at the edge stems from the desire to make high quality computing resources available closer to the users and to reduce the need for end devices to exchange private data with centralized servers \cite{satyanarayanan2017emergence}. There is an ongoing trend to deploy machine learning (ML) algorithms at the edge enabling consumers and corporations to enjoy and explore new opportunities in different application domains, such as automotive, security, surveillance, and other smart city services like healthcare \cite{lim2020federated}. This desire is particularly strong in the healthcare industry where the stakes are high and various stakeholders (including consumers, governments, and service providers) have stressed the need for foolproof safeguards for ensuring data security, user privacy, and ethical data use \cite{ahmad2020developing}. This motivates the use of edge computing in healthcare settings to meet the high expected standards for patient privacy and security as well as stringent requirements for quality of service (high reliability and low latency). 

\begin{figure}[!ht]
\centering
\includegraphics[width=0.4\textwidth]{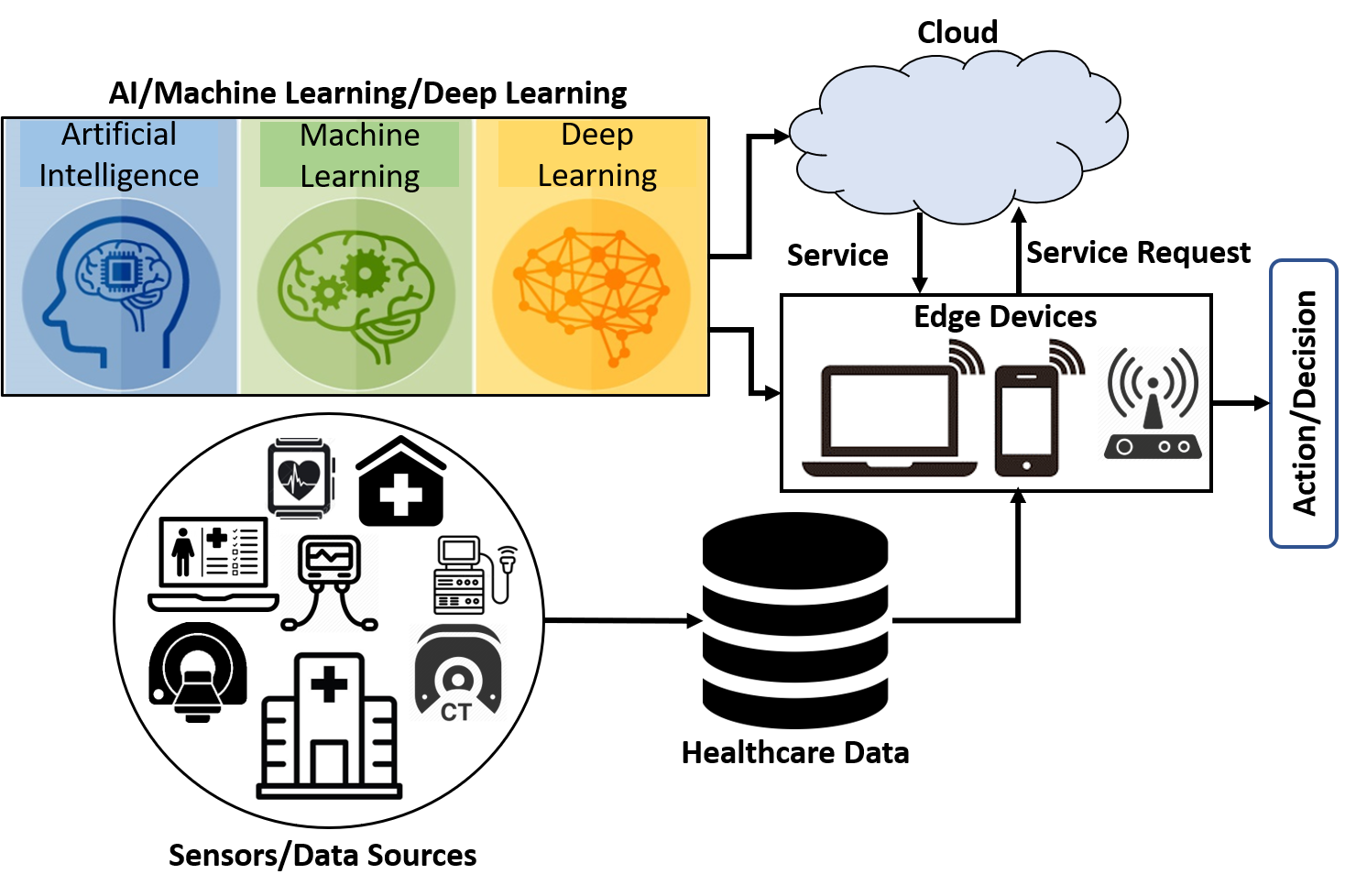}
\caption{An illustration of AI/ML at edge in an IoT empowered healthcare environment.}
	\label{fig:city_edge}
\end{figure}

Healthcare is a major application domain that can benefit from edge computing. Generally, healthcare centers in remote areas lack advanced medical equipment and other healthcare facilities resulting in poorer access to health services by the people living there.  Thanks to the recent advancement in telemedicine, the provision of health services remotely, using audiovisual technology, is a reality now. Large volumes of medical data including ultrasound and X-ray images could be transmitted to major hospitals with advanced diagnosis facilities for diagnostic and training purposes.  However, there are several challenges associated with a typical cloud-based infrastructure, such as low bandwidth, high latency, transmission cost, and increasing concerns about data security and privacy. Edge computing, which aims to store and process data locally or closer to edge devices, on the other hand, results in low latency and increased privacy and data security.

A typical IoT environment for smart healthcare is illustrated in Figure \ref{fig:city_edge} in which data collected by different sensors is processed at the edge for different applications using ML techniques. Once the ML predicts an event, the edge devices triggers an action or request service in the cloud. ML algorithms can also be executed concurrently in the cloud as well as at the edge as shown in the figure. However, with local data storage and processing resources (through cloud computing), such applications enjoy a significant improvement in terms of processing time by avoiding networking congestion. More importantly, real-time processing at the edge improves the performance of delay-sensitive applications, such as healthcare, surveillance and automotive, by avoiding any potential latency or delay occurred during data transmission between end devices and the cloud. In addition, ML at the edge results in increased security and privacy as sending data back and forth from the cloud may lead to security threats. 

In this paper, we leverage the capabilities of edge computing in medicine by analyzing and evaluating the potential of intelligent processing of clinical visual data related to COVID-19 at the edge allowing the remote healthcare centers to benefit from the multi-modal collaborative learning paradigm without sharing any information about the modality of the local data and the data itself. A number of recent research efforts have focused on diagnosing COVID-19 using AI and data science methods \cite{latif2020leveraging}; relatively little work has however focused on using edge AI for COVID-19 diagnosis. 

To this aim, we utilize an emerging concept of clustered federated learning (CFL) and propose a CFL-based collaborative learning framework for an automatic multi-modal diagnosis of COVID-19. Our approach is well suited to the task of COVID-19 diagnosis as visual data (i.e., CT scans, X-rays, and ultrasound) is collected at different centers and could be used to build a joint/shared ML model in a cloud-edge infrastructure able to diagnose COVID-19 in both X-ray and Ultrasound images. The proposed framework is evaluated on two benchmark datasets under different experimental setups and we have achieved encouraging results using CFL that are comparable with the baseline results (when the model is trained with central data). We also discuss in detail the potential applications, associated challenges, technologies, tools, and techniques available for deploying ML at the edge in such privacy and delay-sensitive applications. \textit{We note that we use the term multi-modal model to represent a single model capable of diagnosing COVID-19 in both X-ray and Ultrasound imagery when provided separately}.

\textit{The main contributions of the paper are as follows}:
\begin{enumerate}
    \item To highlight the potential of intelligent processing of clinical data at the edge, we propose a collaborative learning framework for COVID-19 diagnosis by leveraging a CFL approach enabling remote healthcare centers to benefit from each other's data without sharing the data itself and associated information. 
    \item We also demonstrate how the performance of conventional FL is affected by the divergence of the distribution of data from different sources (i.e., X-ray and Ultrasound imagery), and how CFL can help to mitigate the adverse impact.
     \item We also highlight the potential challenges and enabling factors that enable the deployment of ML/DL models to the edge. 
    \item Finally, we elaborate on the open research issues related to deploying ML at the edge for healthcare applications that require further investigation. 
\end{enumerate}

\textit{Organization of the paper:} The rest of the paper is organized as follows. Section \ref{sec:related} provides a broad discussion of the related work on automated COVID-19 diagnosis as well as the different challenges encountered in deploying ML on the edge along with a discussion on enabling technologies. The case study for collaborative learning for multi-modal diagnosis of COVID-19 is presented in Section \ref{sec:case_study} and results are presented in Section \ref{sec:results}. Various open research issues that require further investigation are presented in Section \ref{sec:open}. Finally, Section \ref{conclusion} provides some concluding remarks.

\section{Background}
\label{sec:apps_challenges}

In this section, we provide background on the related work on automated COVID-19 diagnosis as well as a discussion on the potential challenges that hinder deployment ML at the edge along with the different enabling technologies.

\subsection{Existing Automated COVID-19 Diagnosis Work}
\label{sec:related}

COVID-19 has been a strong focus of the research community in 2020, especially after it was declared in March by the World Health Organization (WHO) to be a pandemic, with diverse efforts focusing on diagnosis \cite{tang2020laboratory}, treatment \cite{hendaus2020remdesivir}, and the development of the potential vaccine \cite{hotez2020covid}. Data science methods---particularly, ML and data visualization techniques---are playing a major role in the international response against the COVID-19 pandemic with some key applications being risk assessment, contact tracking, fake news detection, sentiment analysis, and screening and diagnosis \cite{latif2020leveraging}. The focus of this paper is on automated screening and diagnosis; we shall discuss next some of the prominent related techniques relying on different types of information (e.g., audio and visual data) that have been proposed. 

A number of efforts have focused on automated image analysis in a bid to speed up the COVID-19 diagnosis process \cite{ai2020correlation}. To this aim, three different medical imaging modalities, namely computerized tomography (CT), Ultrasound scans, and X-radiation (X-ray), have been mostly exploited. To facilitate research on image-based solutions for COVID-19 diagnosis, several datasets have been collected and made publicly available \cite{ai2020correlation,maghdid2020diagnosing}. For instance, Maghdid et al. \cite{maghdid2020diagnosing} collected a comprehensive dataset containing a total of 170 X-rays and 361 CT scan images from different sources. Cohen et al. \cite{cohen2020covidProspective} also provide a collection of X-rays and CT scans of confirmed COVID-19 patients. A collection of COVID-19 patients' CT scans has also been made publicly available for research purposes in \cite{COVID-19_dataset,zhao2020covid}. Born et al. \cite{born2020pocovid}, on the other hand, provide a lung ultrasound (POCUS) dataset containing a total of 1103 images including 654 COVID-19, 277 bacterial pneumonia, and 277 healthy controls samples. 

A vast majority of the image-based solutions for COVID-19 diagnosis relies on CT scan images. For instance, Wan et al. \cite{wang2020deep} proposed a deep learning model for extracting COVID-19's specific features/textures in CT scans of confirmed cases to extract useful clinical insight before the pathogenic tests. An evaluation of a reasonable amount of confirmed cases showed encouraging results with an average test accuracy of 73.1\%. Butt et al. \cite{butt2020deep} proposed a two-phase solution for COVID-19 diagnosis in CT scans. Initially, a pre-trained 3D Convolutional Neural Network (CNN) is employed to extract potential infectious regions in CT scans followed by a CNN-based classification framework to classify the candidate regions into COVID-19, influenza, and non-infectious regions. Li et al. \cite{li2020artificial} also proposed a 3D CNN-based framework to extract both local and global deep features for diagnosis COVID-19 in CT scans. One of the key challenges to CNN-based solution is the unavailability of a large-scale CT scans datasets. In order to deal with the challenge, Afshar et al. \cite{afshar2020covid} proposed a Capsule Networks based deep learning framework, namely COVID-CAPS, for COVID-19 diagnosis in X-ray images. Moreover, to further enhance the capabilities of the proposed model, the authors used an external dataset composed of 94, 323 frontal view chest X-ray images for pre-training and transfer learning purposes. 

There are also methods relying on X-ray images for COVID-19 diagnosis. For instance, in \cite{wang2020covid} a pre-trained deep model is fine-tuned on X-ray images for COVID-19 diagnosis. Similarly, Sethy et al. \cite{sethy2020detection} trained a Support Vector Machine (SVM) classifier on features extracted via ResNet-50 \cite{he2016deep} from X-ray images for classification of COVID-19 and non-COVID-19 cases. Ali et al. \cite{narin2020automatic} evaluated the performance of several existing deep models in diagnosing COVID-19 in X-ray images. Islam et al. \cite{islam2020combined} on the other hand proposed a deep framework combining CNNs and Recurrent Neural Networks (RNNs) for diagnosis of COVID-19 in X-ray images. Initially, features are extracted with a CNN, which are then feed into a Long short-term memory (LSTM) for diagnosis/detection purposes. Kassani et al. \cite{kassani2020automatic} provide a detailed evaluation of several existing deep models and classification algorithms to find a best combination for COVID-19 diagnosis in both X-ray and CT scans. However, both modalities are treated individually. The deep models are used for feature extraction, which is then fed into different classification algorithms. 

Some image-based COVID-19 diagnosis methods also rely on a recently introduced concept of Federated Learning (FL) to ensure data privacy in a collaborative learning environment, where several hospitals can participate in training a global ML model. For instance, in \cite{xu2020collaborative} a deep model is collaboratively trained in a federated learning environment on CT scans collected from different sources. Kumar et al. \cite{kumar2020blockchain} on the other hand proposed a blockchain-FL-based framework for collecting data (CT scans) from different hospitals, and collaboratively training a global deep model. Moreover, several exiting deep models have also been evaluated in the proposed federated learning framework. In \cite{vaid2020federated}, a federated learning technique is employed for training a global model on electronic health records from various hospitals to predict mortality within seven days in hospitalized COVID-19 patients.

\subsection{Challenges in Deploying ML at the Edge}

\subsubsection{Resource Scarcity and Heterogeneity}
Heterogeneous edge devices with varying computational, storage and communication resources are a major bottleneck for the deployment of ML on the edge. ML algorithms in general and deep learning (DL) in particular require a large amount of computational and processing resources making the deployment of ML impractical in several edge computing applications. DL models are usually large and are computationally expensive, as both the training of a deep model and its inferences are typically performed on power-hungry GPUs and servers while, the edge devices are designed to be operated at low power and usually have frugal memory, therefore, deploying DL models on the edge devices is very challenging. One another important challenge is the availability of a power source at edge device, i.e., a battery with long power backup is always desirable in a typical edge computing network. In addition, the size of the network and systems constraints is also a major challenge that can result in only a few devices being active at a time \cite{li2020federated}.  

\subsubsection{Network Communication}
The heterogeneity of the computational and communication resources also led to slow and unstable communication. In addition to resource heterogeneity, there are other considerations as well, e.g., the Internet upload speed is typically much slower than the download speed \cite{lim2020federated}. Therefore, in an edge computing environment in which ML/DL models are being trained on the client site stable and powerful Internet connection is always desirable, otherwise, the unstable clients will be disconnected from the network that will result in a drop in performance.  On the other hand, deploying ML at the edge saves expensive communication, i.e., we do not require the local (raw) data to be transmitted to the cloud/server.  

\subsubsection{Statistical Heterogeneity}
The statistical heterogeneity, due to data generated by different types of devices in an edge computing  environment, can lead to many efficiency challenges. For instance, the optimization/training of a ML/DL hyperparameters becomes difficult, which directly affect its performance. To address the statistical heterogeneity techniques such meta-learning can be used that can enable device-specific modeling \cite{li2019differentially}. 

\subsubsection{Privacy and Security Challenges}
Despite being able to train joint models with sharing data in a collaborative learning environment using FL, privacy and security challenges arise with the presence of malicious devices. For instance, an adversary can learn sensitive information using the model parameters and the shared model. As shown in \cite{melis2019exploiting}, privacy-related information can be inferred from the shared weights even without getting access to the data itself. 
To restrain leakage of privacy-related information from the shared model, different privacy-preserving techniques can be leveraged, such as cryptographic approaches and differential privacy \cite{qayyum2020secure}.     

\subsubsection{Adversarial ML} 
Despite the state of the art performance of ML/DL techniques in solving complex tasks, these techniques have been found vulnerable to carefully crafted adversarial examples \cite{qayyum2020securing}. In a federated learning setup, a client or multiple clients can be compromised to realize the attacks on the whole network. For instance, local poisoning attacks using compromised attacker devices are presented in \cite{fang2019local}. The authors demonstrated that their proposed attacks can increase the error rates of the distributively trained model on four real-world datasets. Moreover, a systematic review focused on different adversarial ML attacks and defenses for cloud-hosted ML models can be found in \cite{10.3389/fdata.2020.587139}.

\subsection{Enabling Technologies: Building Blocks for ML at Edge} 

\subsubsection{Schemes for deploying ML at the Edge}
\label{AI_edge}

In recent years, enormous growth has been observed in the computational power of edge devices, allowing them to play a more important role than just collecting data in IoTs. ML can contribute significantly in fully utilizing the potential of edge devices in numerous exciting applications (e.g., smart healthcare using wearables technologies and AI-empowered sensors, etc.) and turn them into more useful components of an IoT environment \cite{zhu2020toward}.  ML could be employed at the edge in several ways, such as inference, sensor fusion, transfer learning, generative models, and self-improving devices. In this section, we briefly describe some of the most commonly used schemes.  

\begin{itemize}
    \item \textit{Inference}: The inference capabilities of ML, which aims predicting unseen objects/classes based on the previous knowledge/trained data, help the IoTs to perform different activities, such as cancer prognosis, brain tumor classification, and other clinical data analysis at the edge devices resulting in reduced latency and bandwidth in telemedicine \cite{gobieski2019intelligence}. 
    
    \item \textit{Sensor Fusion}: ML in conjunction with signal processing algorithms can be used for the fusion of information from different sensors enabling efficient utilization of the available information. With fusion capabilities, individual sensors in an IoT environment can be converted into sophisticated synthetic sensors 
    to solve complex problems more accurately. For instance, in healthcare data from several sensors/sources can be combined efficiently to predict a clinical event, such as heart failure \cite{nagpaldeep}.
    
    \item \textit{Transfer Learning}: Transfer learning, which aims to re-utilize the knowledge of one domain in another domain by fine-tuning a pre-trained model trained on a larger dataset, can help them to learn on a smaller dataset with less computational resources.
    In an IoT environment and in particular, in healthcare applications where the data is scarcely available, the transfer learning technique can be used to balance workload and latency where the pre-trained models are put at the cloud and are shared among edge devices to be fine-tuned for specific tasks \cite{zhou2018robust}.   
    
\item \textit{Generative Models}: Generative learning can also be useful in edge computing where generative models can be used for the approximation of the original data at the clouds to be used for training models at edge devices for applications with less training samples or to solve complex tasks with minimal computation from the clouds. Generative deep models have already been explored for the generation of synthetic medical images \cite{qayyum2020single}.
 
 \item \textit{Self-improving devices}: In a typical IoT environment, ML techniques can also be used to enable end devices to optimize their performance and improve continuously based on the collected data and behaviors of other devices. Such strategies help to configure the devices faster which ultimately leads to faster and efficient implementation and deployment.
 
 \end{itemize}

\subsubsection{Hardware Optimization Techniques}
\label{hardware_optimization}
For successful deployment of ML at the edge, the two critical requirements of edge computing---namely (i) low power consumption, and (ii) high performance---need to be fulfilled. Thus, off-the-shelf solutions are not practical to intelligent processing of data at the edge devices, and custom hardware architectures need to be developed. In this section, we discuss some hardware optimization techniques to optimize hardware resources for deploying ML at the edge.    

\paragraph{Decentrailized Distributed Computing}
In edge computing, computations are completely or largely performed on end devices in a distributed computing fashion. Also, edge computing brings data, applications, and services closer to end devices while eliminating the need for centralized clouds that requires infrastructure decentralization, such kind of decentralization can be efficiently achieved using blockchain technologies \cite{booz2017decentralized}. Therefore, computational resources can be shared among end/edge devices by employing blockchain and smart contracts technologies thus allowing computational resources demanding ML applications to be deployed at the edge. For instance, different design requirements and challenges in designing a decentralized edge computing and IoT ecosystem are presented in \cite{psaras2018decentralised}. This study is specifically focused on the need of using decentralized trust schemes for the elimination of trust in centralized entities and highlights the potential of using distributed ledger technology, i.e., blockchain for achieving the feature of decentralization. The backbone of blockchain technologies is the distributed consensus mechanism enabling secure communication among trust-less participants without the intervention of a central controlling unit. There are many facets of blockchain with different distributed consensus methods that can be used for edge-centric IoT systems \cite{zhao2020privacy}. 

\paragraph{AI Co-Processors}
Portable intelligent and dedicated co-processors are considered to be the driving force for deploying AI/ML models at the edge. Different types of specialized processors can be integrated into a single system or chip thus forming a heterogeneous computing paradigm optimized for a specific type of task. In general, AI co-processors have two common features: (1) enables parallel computing using multiple mini-cores; (2) enables accelerated data fetching using distributed memory that is placed right to mini-cores. 

\subsection{Algorithmic Optimization Techniques}
\label{algorithmic_optimization}
The development and advancement of ML algorithms are promising aspects that facilitate the successful application of ML at the edge. In this regard, various algorithms and techniques can be leveraged to enhance and reduce the computation of the parameters in ML models by exploiting different properties such as sparsity. The widely used methods are described below.   

\begin{figure*}[!ht]
\centering
\includegraphics[width=0.9\linewidth]{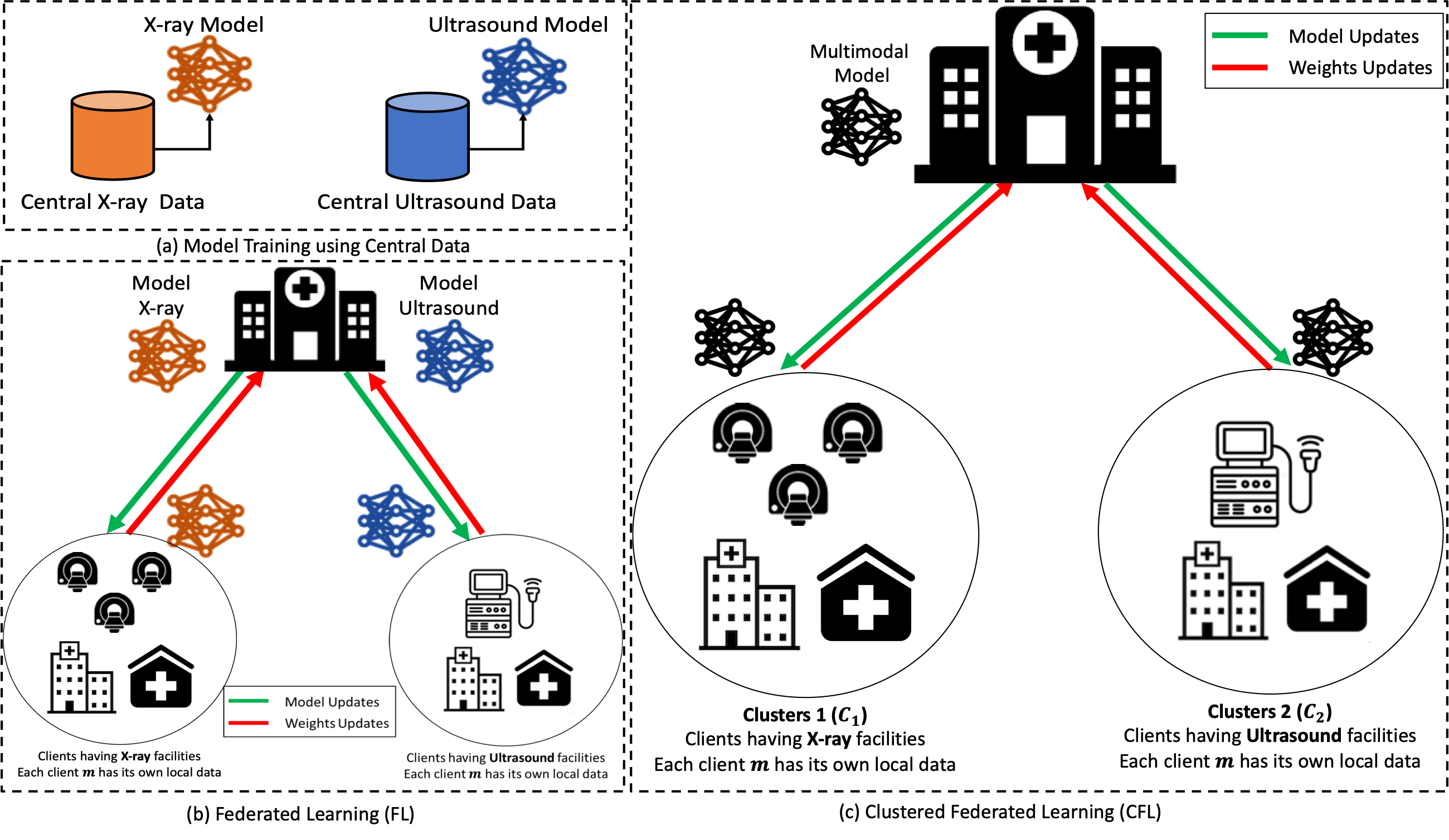}
\caption{The proposed clustered federated learning based collaborative learning paradigm (Fig. \ref{fig:cfl_workflow} (c)) versus the method of model training using central data (Fig. \ref{fig:cfl_workflow} (a)) and the conventional federated learning model training method for multiple modalities (Fig. \ref{fig:cfl_workflow} (b)). The term ``clients" refers to hospitals, clinics, and medical imaging facilities.}
	\label{fig:cfl_workflow}
\end{figure*}

\begin{itemize}
    \item \textit{Parameter Efficient Networks:} To efficiently deploy ML models at the edge, computation and memory-efficient architectures of ML/DL models are highly desirable. To facilitate embedded ML computing, various architectures of ML models have been proposed in the literature that can be leveraged to deploy ML models on the edge, e.g., Mobile Net \cite{howard2017mobilenets} and SqueezeNet \cite{iandola2016squeezenet}. These architectures are designed with a key focus on reducing computation costs associated with the training and inferences of ML models while maintaining accuracy. 

    \item  \textit{Network Pruning:} 
     The literature suggests that a penalty of neurons in the trained model does not contribute towards the final accuracy, therefore, such neurons can be prune to save some memory. Google's Learn2Compress\footnote{https://ai.googleblog.com/2018/05/custom-on-device-ml-models.html} has found that neurons can be reduced by a factor of 2 while retaining an overall accuracy of 97\%. To this aim, several algorithms have been proposed in the literature, such as learning important connections and weights among neurons \cite{han2015learning} and learning structural sparsity in deep models \cite{wen2016learning}. Moreover, many ML models perform parameter computation using 32-bit float values. On the other hand, edge devices typically operate on 8-bit values or less. Therefore, the model size can be significantly reduced by reducing precision. 

    \item \textit{Network Distillation:} Network distillation is a method for transferring knowledge learned by a larger model to a smaller model. Together with transfer learning, which deals with the transfer of knowledge learned from one domain to another domain, network distillation holds the substantial potential to significantly reduce model size without comprising much on performances in terms of accuracy. In addition, network distillation can be benefited from other hyperparameters tuning algorithms as well. For instance, the distillation method has been successfully used for application-specific and resource-constrained IoT platforms \cite{yim2017gift}.  
\end{itemize}

\section{COVID-19 Diagnosis using Collaborative Federated Learning}
\label{sec:case_study}

In this section, we consider the problem of developing a single ML model for classification of chest images from multiple sources (such as X-rays and Ultrasound). Consider a clustered federated learning (CFL) setup as shown in Figure \ref{fig:cfl_workflow} resembles the actual federated learning settings \cite{sattler2019clustered}. Clients in each cluster represent the healthcare entities (remote medical imaging facilities) and major hospitals or other government entities (e.g., ministry of health) play the role of the cloud server facilitating the weights aggregation and updates. The problem formulation for collaborative learning is described below.  
 
\subsection{Problem Formulation}
In this task, we are interested in learning a shared model $M_s$ in a collaborative fashion using clustered federated learning (CFL). As shown in Figure \ref{fig:cfl_workflow}, there are two clusters each having different kind of imaging modality, i.e., cluster 1 ($C_1$) has clients having X-ray imagining facility and clients in cluster 2 ($C_2$) posses ultrasound imagining facilities, therefore, each cluster $C_k$ is disjoint and has different data distribution $\mathcal{D}_k$. Each client $m$ in cluster $C_k$ has drawn its samples $z^{k,1},\dots,z^{k,m}$ from the distribution $\mathcal{D}_k$ such that there are no overlapping samples among the clients. We have formulated the problem of collaborative learning as supervised learning problem such that each sample $z^{k,m}$ contains a pair of data sample $x^{k,m}$ and its corresponding class label $y^{k,m}$, denoted by $z^{k,m} = (x^{k,m},y^{k,m})$.  Furthermore, we assume that each client does not have any knowledge either about the identity and data of every other client within the same cluster as well as in the other cluster. The major hospital (aka server) shares a shared model $M_s$ and initial weights $W_0$ with each client of every cluster. After receiving the $M_s$ and $W_0$, each client trains the shared model (i.e., $M_s$) using its own local data $D_{k,m}$, where $k=\{1,2\}$ and $m$ denotes the number of clients in each cluster $C_k$. After that, every client in each cluster shares the learned weights $W_{r,k,m}$ to the server, where $r$ denotes the communication round/iteration number. After receiving the weight updates from each client, the server performs federated averaging using Eq. \ref{eq:fed_avg}.

\begin{equation}
\label{eq:fed_avg}
    W_{r} = \frac{1}{n} \sum_i^n w_i \times W_{i}
\end{equation}
 
Where, $n$ denotes the total number of clients participating in the CFL setup (i.e., $n=|C_1|+|C_2|$) and $w$ is a weighting factor that specifies the weight-age given to the weights of each client. Then the server updates the new weights (i.e., update its copy of $M_s$ with $W_{r}$) and performs the inference using its multi-modal test data (the two modalities, i.e., X-ray and Ultrasound are merged to make the testing data multi-modal). After testing the performance of $M_s$ at the communication round $r$, the server shares the updated weights $W_{r}$ with all clients in each cluster and repeats the process until the specified criteria or desired performance is achieved. The algorithm for collaborative multi-modal learning using CFL is presented in Algorithm \ref{algo}.

\begin{algorithm}[H]
\textbf{Input:} Shared Model $M_s$, Clusters $k$, Initial Model Weights $W_0$, Set of Clients $m$, Communication Rounds $R$, Epochs $E$, Batch Size $B$, and Learning Rate $\eta$ \\
\textbf{Output:}  Updated Weights $W_{r}$ \\
\textbf{Initialize:} $W_0$, $R$, $E$, $B$, and $\eta$
  \For{$r = 1,..., R$}{ 
  \For {$i = 1,..., k$ \textit{\textbf{in parallel}}}{ 
  \For {$j = 1,..., m$ \textit{\textbf{in parallel}}}{ 
  \eIf{$r==1$}{
   $W_{i,m} \leftarrow W_0$\;
   $M_s \leftarrow W_0$\;
   \For {$e \in E$}{
   Using $B$ \& $\eta$\;
   \textbf{\textit{Train}} $M_s$ using $z^{i,j} = (x^{i,j},y^{i,j})$\;
    Get $W_{j}$ from $M_s$\;
   }
   }{
   $M_s \leftarrow W_r$\; 
   \For {$e \in E$}{
   Using $B$ \& $\eta$\;
   \textbf{\textit{Train}} $M_s$ using $z^{i,j} = (x^{i,j},y^{i,j})$\;
   Get $W_{j}$ from $M_s$\;
   }
  }
  }
  $W_{i,r} = \frac{1}{m} \sum_{j=1}^m w_j \times W_{j}$\;
  }
  \textbf{Return:} $W_r = \frac{1}{k} \sum_{i=1}^k W_{i,r}$\; } \caption{Collaborative learning from data of different sources and modality using CFL}\label{algo}
\end{algorithm}

\subsection{Experimental Setup} 

\subsubsection{Data Description}
For this study, two datasets from different sources one containing chest X-ray \cite{cohen2020covidProspective} and chest ultrasound images \cite{born2020pocovid}, are used. We formulated the problem as binary classification, i.e., differentiating between COVID-19 chest images and normal chest images. Each dataset is divided into two parts, i.e., a training set and a testing set using a split of 80\% and 20\%, respectively. The training portion (i.e., 80\%) of each dataset is further divided into different parts, depending upon the number of clients in that cluster. The distribution of training and testing data of X-ray and Ultrasound datasets over different classes is shown in Table \ref{tab:data_dis}. Moreover, the testing sets from both datasets are merged to develop a joint  testing set that will be used by the server for the evaluation of the performance of a shared model that is being trained in a collaborative fashion using CFL. 

\begin{figure}[!ht]
\centering
\includegraphics[width=0.4\textwidth]{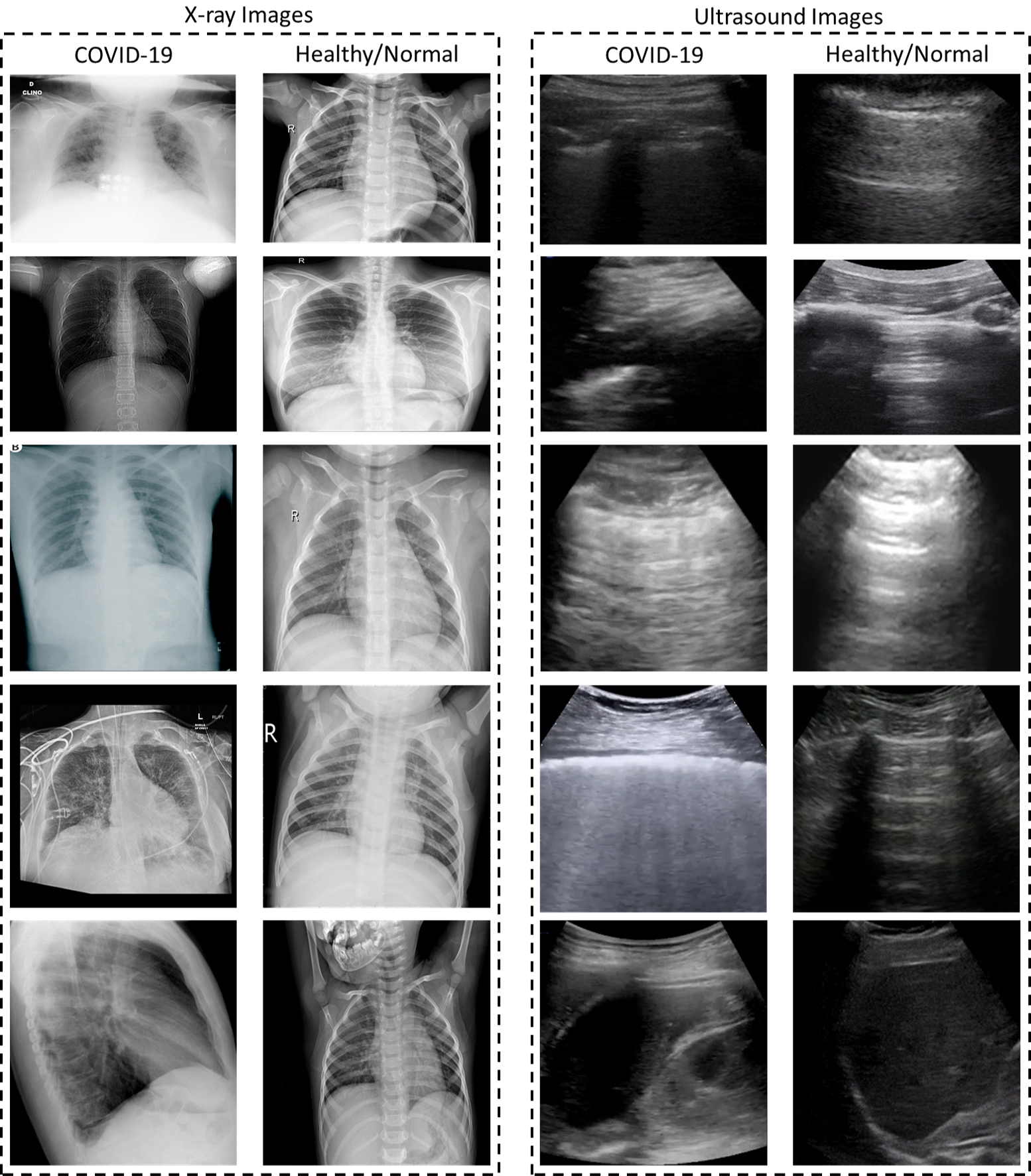}
\caption{The depiction of inter and intra class variations observed in COVID-19 datasets (X-ray \cite{cohen2020covidProspective} and Ultrasound \cite{born2020pocovid}).}
	\label{fig:data_ex}
\end{figure}

We further note that the datasets used in this study have inter and intra class variability in terms of image size and quality, contrast and brightness level, and positioning of subjects, an example is shown in Figure \ref{fig:data_ex}. This is not surprising as these publicly available databases are not standard datasets for COVID-19 detection, and have been curated from different sources \cite{horry2020covid}. Moreover, it is evident from Table \ref{tab:data_dis} that these datasets are highly imbalanced. These limitations make the training of a generalized model more difficult.

\begin{table}[!ht]
\centering
\caption{The distribution of training and testing data of X-ray and Ultrasound datasets over different classes.}
\begin{tabular}{|l|l|c|c|}
\hline
\multicolumn{1}{|c|}{Data} & \multicolumn{1}{c|}{Class} & Training Data (80\%) & Test Data (20\%) \\ \hline
\multirow{3}{*}{X-ray} & COVID-19 & 179 & 44 \\ \cline{2-4} 
 & Healthy & 1072 & 269 \\ \hline
\multirow{3}{*}{Ultrasound} & COVID-19 & 319 & 80 \\ \cline{2-4} 
 & Healthy & 116 & 30 \\ \hline
\end{tabular}
\label{tab:data_dis}
\end{table}

\subsubsection{Model Architecture and Implementation Details}
In our experiments, we have used the VGG16 model with one extra convolutional layer and three fully connected layers stacked before its original output layer. Each image is first converted into a gray-scale image, which is then resized to a dimension of $256 \times 256$. Moreover, the resized images are normalized before feeding into the model. The model is trained using \textit{Adam} optimizer with a learning rate of $0.0001$ at each client. We use different types of standard data augmentation techniques for training the models. Furthermore, to address the problem of imbalanced classes, we propose to use focal loss \cite{lin2017focal}, which is suited for such issues in binary classification tasks. The focal loss adds a modulating factor $(1 - p_t)^\gamma$ to the standard cross-entropy loss, where $\gamma \ge 0$ is a tunable focusing parameter. The $\alpha$-balanced variant of focal loss is defined in Eq. \ref{eq:fl}, where $\alpha$ balances the importance of positive/negative examples \cite{lin2017focal}.

\begin{equation}
    FL(p_t) = - \alpha_t (1 - p_t)^\gamma \log(p_t)
\label{eq:fl}
\end{equation}

We note that for the implementation of the proposed work we used \textit{TensorFlow} ML library, and all experiments are performed in a simulated environment. The results of the different experiments are described in the next section.   

\section{Results and Discussions}
\label{sec:results}
In order to show the effectiveness of the proposed multi-modal collaborative learning framework for COVID-19 diagnosis, we performed several experiments. On one side, we aim to evaluate and compare the performances of CFL against two baselines, namely (i) \textit{specialized FL baseline}, and the (ii) \textit{multi-modal\footnote{By the term multi-modal we mean images acquired using different imagining techniques, i.e., modalities (e.g., X-ray and Ultrasound).} conventional FL}. Since CFL aims to tackle the convergence issues of conventional FL schemes due to the diverse distribution of the data, the two baselines, we believe, are appropriate options as a comparison benchmark instead of the state of the art methods for COVID-19 diagnosis. We note that due to the limitations of the dataset, we only consider the divergence in distribution of the data in terms of the nature of the data (i.e., the distribution of ultrasound and X-ray images is different). The first baseline shows the best-case scenario, where separated models for each type of imagery, which we termed as specialized models, are trained in a FL environment. The individual models are trained on X-ray and Ultrasound images with a learning rate of $0.0001$ and a batch size of $32$ resulting into two separate models one for each modality (i.e., X-ray and Ultrasound). The second baseline represents the experimental setup of a conventional FL environment, where the data is distributed among different clients, and a shared ML model is built in a federated environment. The parameters used in different experiments can be found in Table \ref{tab:pars}. 

\begin{table}[!ht]
\centering
\caption{Parameters of clustered federated learning (CFL) experiments.}
\begin{tabular}{|l|c|}
\hline
\textbf{Parameter (s)} & \textbf{Value (s)} \\ \hline
Communication Rounds   &    30, 50, \& 100                \\ \hline
Epochs                 &    5 \& 10           \\ \hline
Batch Size             &    16 \& 32                \\ \hline
Learning Rate          &    $1e^{-3}$               \\ \hline
\end{tabular}
\label{tab:pars}
\end{table}

\begin{table*}[!ht]
\centering
\caption{\MakeUppercase{Comparison against the two baselines in terms of precision, recall, and F1-Score.} \textit{Promising results are obtained by CFL, outperforming the \textit{conventional FL} while slightly lower performance is obtained compared to \textit{Central Baseline} with the added advantage of improved privacy and data security.} $^*$A separate model is trained in federated learning settings for each modality.}
\begin{tabular}{|l|l|c|c|c|c|c|c|c|c|c|}
\hline
\multirow{2}{*}{\textbf{Dataset}}    & \multirow{2}{*}{\textbf{Class}} & \multicolumn{3}{c|}{\textbf{Federated Learning (Specialized$^*$)}}                                                                          & \multicolumn{3}{c|}{\textbf{Federated Learning (multi-modal)}}                                                                   & \multicolumn{3}{c|}{\textbf{Clustered FL (multi-modal)}}                                                        \\ \cline{3-11} 
                                     &                                 & \multicolumn{1}{l|}{\textbf{Precision}} & \multicolumn{1}{l|}{\textbf{Recall}} & \multicolumn{1}{l|}{\textbf{F1-Score}} & \multicolumn{1}{l|}{\textbf{Precision}} & \multicolumn{1}{l|}{\textbf{Recall}} & \multicolumn{1}{l|}{\textbf{F1-Score}} & \multicolumn{1}{l|}{\textbf{Precision}} & \multicolumn{1}{l|}{\textbf{Recall}} & \multicolumn{1}{l|}{\textbf{F1-Score}} \\ \hline
\multirow{2}{*}{\textbf{X-ray}}      & COVID-19                        & 0.73   &   0.82    &  0.77                                  & 0.30                                    & 0.68                                 & 0.41                                   & 0.71                                    & 0.82                                 & 0.76                                   \\ \cline{2-11} 
                                     & Healthy                         & 0.97  &    0.95 &     0.96                                   & 0.93                                    & 0.74                                 & 0.82                                   & 0.97                                    & 0.94                                 & 0.96                                   \\ \hline 
\multirow{2}{*}{\textbf{Ultrasound}} & COVID-19                        & 0.97                                    & 0.95                                 & 0.97                                  & 0.94                                    & 0.76                                 & 0.84                                   & 0.93                                    & 0.95                                 & 0.94                                   \\ \cline{2-11} 
                                     & Healthy                         & 0.88   &   0.93 &     0.90                                & 0.58                                    & 0.87                                 & 0.69                                   & 0.86                                    & 0.80                                 & 0.83                                   \\ \hline
\end{tabular}
\label{tab:pe}
\end{table*}

\begin{figure*}[!ht]
    \centering
    \subfigure[X-ray]{\includegraphics[width=0.4\textwidth]{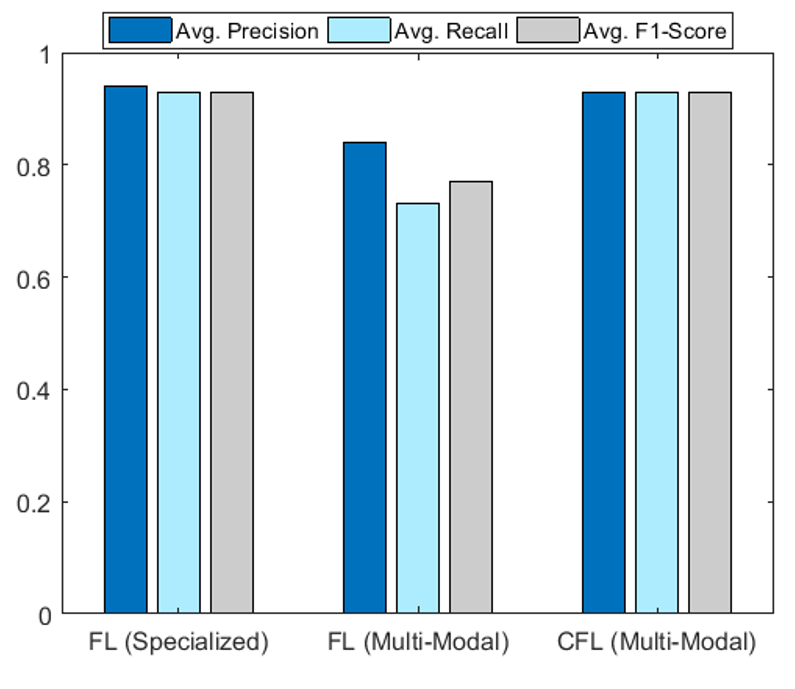}\label{fig:xb}}
    \subfigure[Ultrasound]{\includegraphics[width=0.4\textwidth]{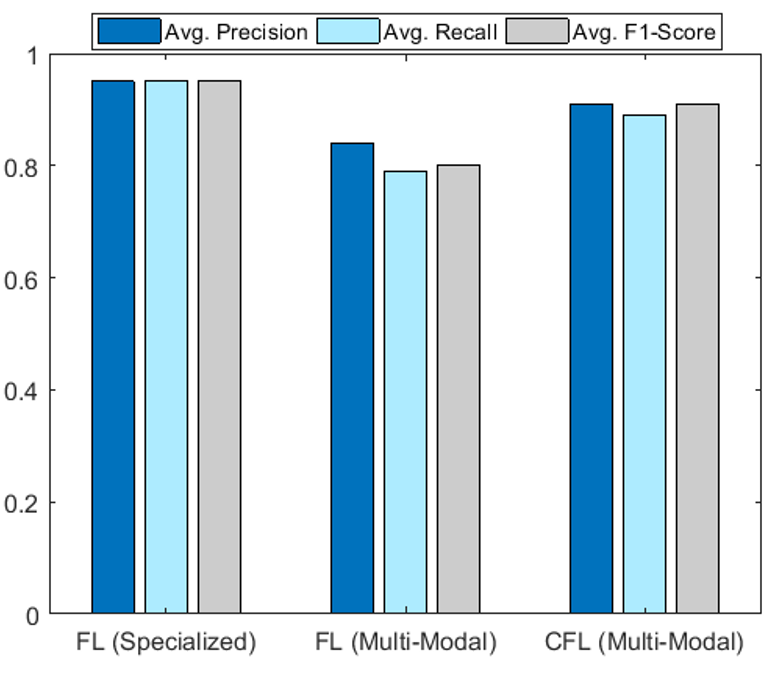}\label{fig:ub}}
    \caption{Comparison of clustered federated learning (CFL) with two baselines (i.e., the specialized models (trained with conventional FL independently for each modality) and conventional federated learning (when the model is trained using multi-modal data)) in terms of average values of precision, recall, and F1-score on X-ray and Ultrasound imagery. }
    \label{fig:avg_comp}
\end{figure*}

Table \ref{tab:pe} and Figure \ref{fig:avg_comp} provides the experimental results per class and overall (per dataset) results, respectively, in terms of precision, recall, and F1-Score. Since the data set is not balanced, so we believe, alone accuracy is not enough to evaluate the proposed method. 
For performance evaluation of the three experimental setups (i.e., the two baselines and CFL), we kept the similar experimental setup where we first train the baseline models with a batch size of 16 (for each modality) and then we train the same model in CFL fashion (i.e., using multi-modal settings) with 5 epochs of local training with a batch size of 16. Then we evaluated the collaboratively trained model with the test data from each cluster (modality), i.e., X-ray and Ultrasound. As can be seen in the Figure \ref{fig:avg_comp}, overall comparable results are observed for multi-modal model trained using CFL compared with the specialized two models trained in a conventional FL environment using X-ray and Ultrasound imagery separately. On the other hand, we can see that CFL performance is considerably better than the performance of multi-modal model trained in a conventional federated learning environment. Moreover, it is evident from the figure that a collaboratively trained model is capable of recognizing the test of images from different modalities without having explicit knowledge about these modalities. Moreover, overall better results are obtained on ultrasound images (Figure \ref{fig:ub}) compared to X-ray imagery (Figure \ref{fig:xb}) for all models.

\begin{figure*}[!ht]
    \centering
    \subfigure[]{\includegraphics[width=0.41\textwidth]{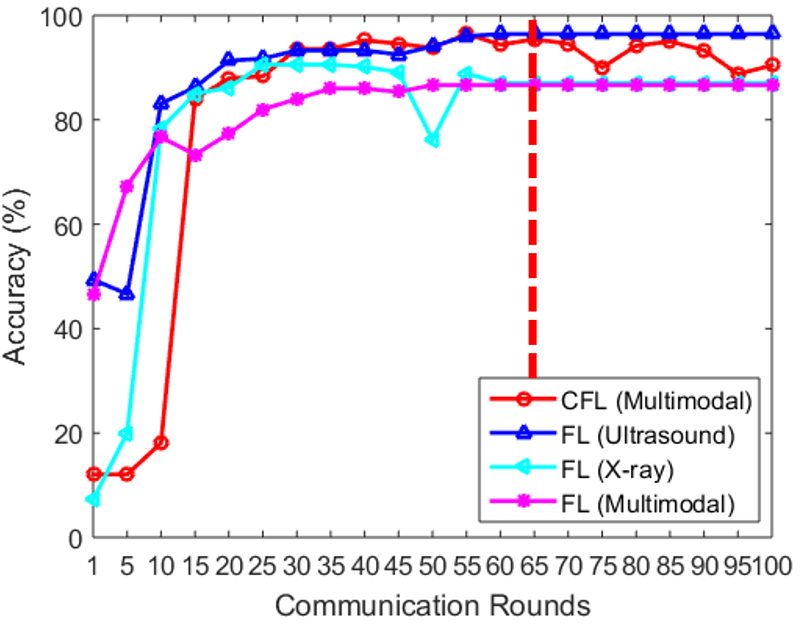}\label{fig:acc_bc}}
    \subfigure[]{\includegraphics[width=0.41\textwidth]{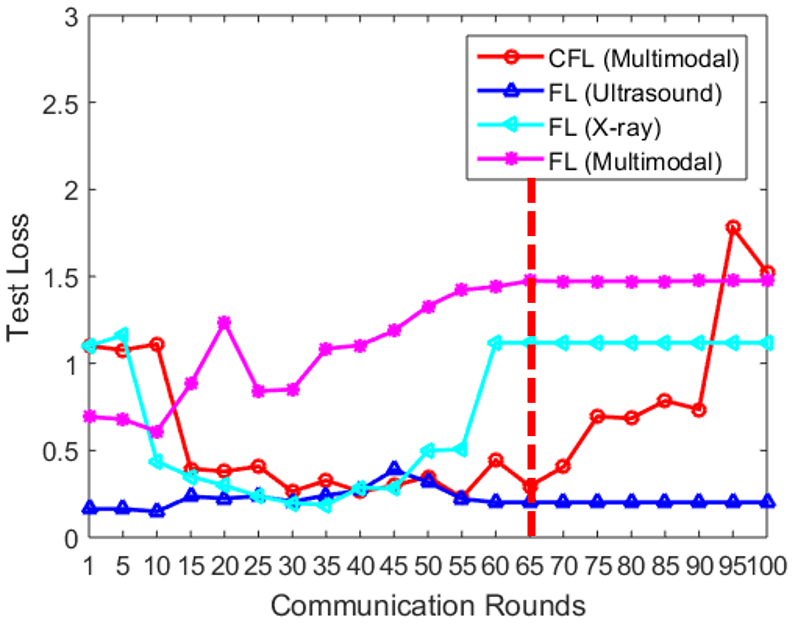}\label{fig:loss_bc}}
    \caption{Comparison of clustered federated learning (CFL) with the specialized models (trained with conventional FL independently for each modality, i.e., X-ray and Ultrasound) and conventional federated learning (when the model is trained using multi-modal data) over increasing number of communication rounds.}
    \label{fig:baseline_comp}
\end{figure*}

In Figure \ref{fig:baseline_comp}, we provide the comparison of the three experimental setups (i.e., specialized models trained in conventional FL settings, multi-modal models trained in a conventional FL and CFL environments) in terms of accuracy and loss at different communication rounds. The figure depicts that the proposed CFL model (which is trained using multi-modal data) provides comparable performance with that of specialized FL models (that are separately trained for each modality). Moreover, it is also evident from the figure that the model trained using multi-modal data in conventional FL settings gets over-fitted after 50 communication rounds. On the counter side, the model keeps on learning in CFL setting, though it also tends to show over-fitting behavior  at later stage communication round as evident in the Figure \ref{fig:baseline_comp}. The vertical red line shown on Figures \ref{fig:acc_bc} and \ref{fig:loss_bc} shows the inflection point beyond which the parameters of the specialized machine learning models of the two clusters (i.e., X-ray and ultrasound) start to diverge from each other. This diversion limits the extent to which the multimodal model can be generalized to fit the underlying multimodal data (i.e., X-ray and ultrasound). Therefore, Figure \ref{fig:loss_bc} provides the insight that the federated learning rounds should be stopped as soon as the inflection point in the value of the loss function is reached. This inflection point identified the number of rounds beyond which the multimodal machine learning model cannot be enhanced.

\subsection{Lessons Learned}

Some key lessons learned from the literature and experiments conducted in this work are:

\begin{itemize}

\item CFL ensures the privacy of the user's local data, as it does not need to be shared with the server for central training. 

\item The communication payload of model weights is far less than the payload of sharing actual data, therefore, it saves bandwidth and as well as time. 

\item It enables the collaborative learning of multi-modal features by shared model $M_s$ without sharing any explicit information about the modality of the local data and the data itself. 

\item More importantly, compared to the conventional FL, CFL ensures better performance in presence of divergence in the data distribution. The divergence in the distribution could be in terms of the distribution of negative and positive samples per class as well as in terms of the nature of data samples as detailed earlier.

\item This particular use-case demonstrates the potential of the method in medical applications where remote smaller healthcare units can benefit from this collaborative learning method. 

\end{itemize}

However, despite these benefits, there are some challenges and limitations as well, e.g., efficiency, security issues, and the optimization of CFL parameters is difficult. Moreover, there is a trade-off in the model performance when we compare model trained using central data and model trained with distributed data using federate learning. In addition, for multi-modal distributively dispersed data, the development of personalized models that are tailored to these modalities is required for local training, which will enhance the efficiency of the shared model and as well as of the models on the client-side. For instance, from our experiments we have learned that the performance of the model being trained in CFL settings start degrading after a particular point (i.e., communication round). Thus highlighting the need of early stopping and the development of optimal stopping criteria except the maximum allowed communication rounds.      

\section{Open Research Issues}
\label{sec:open}
\subsection{Developing Personalized Approaches}
The edge computing network is potentially more heterogeneous as compared to any other central network and clients in an edge computing network vary due to data acquisition resources, network, computational, and storage resources, etc. Moreover, as discussed above in the paper, clients can significantly vary due to statistical heterogeneity, which is usually a great challenge in realistic settings. For example, as we discussed in the above section, developing a multi-modal collaborative learning framework for COVID-19 diagnosis has efficiency challenges due to the aforementioned heterogeneity issues. Therefore, to handle such heterogeneities, the development of personalized and client-specific ML/DL approaches is required. 

\subsection{Adversarialy Robust ML} 
The edge computing network is more prone to security threats, as edge computing network is an ideal environment for adversaries that aim to get desired outcomes or incentives for breaching the network security and privacy of participating agents. This phenomenon becomes, even more, worse with the integration of ML/DL models that are vulnerable to adversarial attacks, which have been already shown effective for healthcare applications \cite{finlayson2019adversarial}. For instance, an adversarial attack on CT scanners in an actual hospital environment by manipulating the hospital's network has already been realized in the literature \cite{mirsky2019ct} and threats of adversarial ML for ML and IoT empowered COVID-19 detection systems are highlighted in \cite{rahman2020adversarial}. To restrain the adversarial attacks, different defensive techniques have been proposed in the literature. However, the adversarially robust methods developed so far are attack specific, i.e., they only work for particular attacks for which they were developed and fail to withstand unforeseen attacks. Therefore, the development of adversarially robust ML/DL models is still an open research problem that demands a proportionate amount of interest from the community with the advancement of ML/DL techniques. Moreover, for the successful deployment of ML/DL models on the edge, in particular, for developing robust healthcare applications, the development of adversarially robust models are of utmost importance.

\subsection{Asynchronous Distributed ML} 
In distributed computing, two approaches are widely used for communication, i.e., synchronous and asynchronous. These approaches are ideal for scenarios where data is instantly available for instance in the central picture archiving and communication system (PACS) of a hospital. However, in realistic settings, the data collection or acquisition might get delayed due to any reason, such as due to some network issue or unavailability of I/O device, etc. Moreover, it is possible that the client (i.e., a small healthcare entity) in an ML-based collaborative computing network is not active at the current iteration/communication round due to some inherent issue, this will result in a delay in the federated parameters update process and will eventually affect the system's overall performance. Therefore, it is worth studying and developing asynchronous approaches for facilitating shared model training for healthcare applications using distributed data.  

\section{Conclusions}
\label{conclusion}

This article provides insights on how edge computing and machine learning advances can be used to provide a solution for COVID-19 diagnosis in an efficient privacy-aware manner thereby allowing remote healthcare units to benefit from collaborative learning paradigm without sharing local data. In particular, we propose using a clustered federated learning (CFL)-based collaborative learning framework to intelligently process visual data at the edge by training a multi-modal ML model capable of diagnoses COVID-19 in both X-ray and Ultrasound imagery. Compared to the conventional FL, CFL is found to better cope with the divergence in distribution of data from different sources (i.e., X-ray and Ultrasound imagery). In the current implementation, we consider the divergence in distribution due to the sources and nature of the data due to the limitations of the datasets. In the future, we will explore how CFL performs in the presence of variances in the distribution of the data in terms of the number of samples per client.

\bibliographystyle{IEEEtran}

\begin{thebibliography}{100}

\bibitem{ahmad2020developing}
Kashif Ahmad, Majdi Maabreh, Mohamed Ghaly, Khalil Khan, Junaid Qadir, and Ala
  Al-Fuqaha.
\newblock Developing future human-centered smart cities: Critical analysis of
  smart city security, interpretability, and ethical challenges.
\newblock {\em arXiv preprint arXiv:2012.09110}, 2020.

\bibitem{lim2020federated}
Wei Yang~Bryan Lim, Nguyen~Cong Luong, Dinh~Thai Hoang, Yutao Jiao, Ying-Chang
  Liang, Qiang Yang, Dusit Niyato, and Chunyan Miao.
\newblock Federated learning in mobile edge networks: A comprehensive survey.
\newblock {\em IEEE Communications Surveys \& Tutorials}, 2020.

\bibitem{latif2020leveraging}
S.~{Latif}, M.~{Usman}, S.~{Manzoor}, W.~{Iqbal}, J.~{Qadir}, G.~{Tyson},
  I.~{Castro}, A.~{Razi}, M.~N.~K. {Boulos}, A.~{Weller}, and J.~{Crowcroft}.
\newblock Leveraging data science to combat covid-19: A comprehensive review.
\newblock {\em IEEE Transactions on Artificial Intelligence}, 1(1):85--103,
  2020.

\bibitem{tang2020laboratory}
Yi-Wei Tang, Jonathan~E Schmitz, David~H Persing, and Charles~W Stratton.
\newblock Laboratory diagnosis of covid-19: current issues and challenges.
\newblock {\em Journal of clinical microbiology}, 58(6), 2020.

\bibitem{hendaus2020remdesivir}
Mohamed~A Hendaus.
\newblock Remdesivir in the treatment of coronavirus disease 2019 (covid-19): A
  simplified summary.
\newblock {\em Journal of Biomolecular Structure and Dynamics}, 2020.

\bibitem{hotez2020covid}
Peter~J Hotez, David~B Corry, and Maria~Elena Bottazzi.
\newblock {COVID-19 vaccine design: the Janus face of immune enhancement}.
\newblock {\em Nature Reviews Immunology}, 20(6):347--348, 2020.

\bibitem{ai2020correlation}
Tao Ai, Zhenlu Yang, Hongyan Hou, Chenao Zhan, Chong Chen, Wenzhi Lv, Qian Tao,
  Ziyong Sun, and Liming Xia.
\newblock Correlation of chest ct and rt-pcr testing in coronavirus disease
  2019 (covid-19) in china: a report of 1014 cases.
\newblock {\em Radiology}, page 200642, 2020.

\bibitem{kalkreuth2020covid}
Roman Kalkreuth and Paul Kaufmann.
\newblock {COVID-19: a survey on public medical imaging data resources}.
\newblock {\em arXiv preprint arXiv:2004.04569}, 2020.

\bibitem{maghdid2020diagnosing}
Halgurd~S Maghdid, Aras~T Asaad, Kayhan~Zrar Ghafoor, Ali~Safaa Sadiq, and
  Muhammad~Khurram Khan.
\newblock Diagnosing {COVID-19 pneumonia from X-ray and CT} images using deep
  learning and transfer learning algorithms.
\newblock {\em arXiv preprint arXiv:2004.00038}, 2020.

\bibitem{cohen2020covidProspective}
Joseph~Paul Cohen, Paul Morrison, Lan Dao, Karsten Roth, Tim~Q Duong, and
  Marzyeh Ghassemi.
\newblock {COVID-19 Image Data Collection: Prospective Predictions Are the
  Future}.
\newblock {\em arXiv 2006.11988}, 2020.

\bibitem{COVID-19_dataset}
{COVID-19 CT segmentation dataset}.
\newblock \url{http://medicalsegmentation.com/covid19/}.
\newblock Accessed: 2020-08-127.

\bibitem{zhao2020covid}
Jinyu Zhao, Yichen Zhang, Xuehai He, and Pengtao Xie.
\newblock Covid-ct-dataset: a ct scan dataset about covid-19.
\newblock {\em arXiv preprint arXiv:2003.13865}, 2020.

\bibitem{born2020pocovid}
Jannis Born, Gabriel Br{\"a}ndle, Manuel Cossio, Marion Disdier, Julie Goulet,
  J{\'e}r{\'e}mie Roulin, and Nina Wiedemann.
\newblock {POCOVID-Net: automatic detection of COVID-19 from a new lung
  ultrasound imaging dataset (POCUS)}.
\newblock {\em arXiv preprint arXiv:2004.12084}, 2020.

\bibitem{wang2020deep}
Shuai Wang, Bo~Kang, Jinlu Ma, Xianjun Zeng, Mingming Xiao, Jia Guo, Mengjiao
  Cai, Jingyi Yang, Yaodong Li, Xiangfei Meng, et~al.
\newblock A deep learning algorithm using ct images to screen for corona virus
  disease (covid-19).
\newblock {\em MedRxiv}, 2020.

\bibitem{butt2020deep}
Charmaine Butt, Jagpal Gill, David Chun, and Benson~A Babu.
\newblock Deep learning system to screen coronavirus disease 2019 pneumonia.
\newblock {\em Applied Intelligence}, page~1, 2020.

\bibitem{li2020artificial}
Lin Li, Lixin Qin, Zeguo Xu, Youbing Yin, Xin Wang, Bin Kong, Junjie Bai,
  Yi~Lu, Zhenghan Fang, Qi~Song, et~al.
\newblock Artificial intelligence distinguishes {COVID-19 from community
  acquired pneumonia on chest CT}.
\newblock {\em Radiology}, 2020.

\bibitem{afshar2020covid}
Parnian Afshar, Shahin Heidarian, Farnoosh Naderkhani, Anastasia Oikonomou,
  Konstantinos~N Plataniotis, and Arash Mohammadi.
\newblock {COVID-CAPS: A capsule network-based framework for identification of
  COVID-19 cases from X-ray images}.
\newblock {\em arXiv preprint arXiv:2004.02696}, 2020.

\bibitem{wang2020covid}
Linda Wang and Alexander Wong.
\newblock Covid-net: A tailored deep convolutional neural network design for
  detection of {COVID-19} cases from chest {X-Ray} images.
\newblock {\em arXiv preprint arXiv:2003.09871}, 2020.

\bibitem{sethy2020detection}
Prabira~Kumar Sethy and Santi~Kumari Behera.
\newblock Detection of coronavirus disease (covid-19) based on deep features.
\newblock {\em Preprints}, 2020030300:2020, 2020.

\bibitem{he2016deep}
Kaiming He, Xiangyu Zhang, Shaoqing Ren, and Jian Sun.
\newblock Deep residual learning for image recognition.
\newblock In {\em Proceedings of the IEEE conference on computer vision and
  pattern recognition}, pages 770--778, 2016.

\bibitem{narin2020automatic}
Ali Narin, Ceren Kaya, and Ziynet Pamuk.
\newblock Automatic detection of coronavirus disease (covid-19) using x-ray
  images and deep convolutional neural networks.
\newblock {\em arXiv preprint arXiv:2003.10849}, 2020.

\bibitem{islam2020combined}
Md~Zabirul Islam, Md~Milon Islam, and Amanullah Asraf.
\newblock A combined deep cnn-lstm network for the detection of novel
  coronavirus (covid-19) using x-ray images.
\newblock {\em Informatics in Medicine Unlocked}, page 100412, 2020.

\bibitem{kassani2020automatic}
Sara~Hosseinzadeh Kassani, Peyman~Hosseinzadeh Kassasni, Michal~J Wesolowski,
  Kevin~A Schneider, and Ralph Deters.
\newblock Automatic detection of coronavirus disease (covid-19) in x-ray and ct
  images: A machine learning-based approach.
\newblock {\em arXiv preprint arXiv:2004.10641}, 2020.

\bibitem{xu2020collaborative}
Yongchao Xu, Liya Ma, Fan Yang, Yanyan Chen, Ke~Ma, Jiehua Yang, Xian Yang,
  Yaobing Chen, Chang Shu, Ziwei Fan, et~al.
\newblock {A collaborative online AI engine for CT-based COVID-19 diagnosis}.
\newblock {\em medRxiv}, 2020.

\bibitem{kumar2020blockchain}
Rajesh Kumar, Abdullah~Aman Khan, Sinmin Zhang, WenYong Wang, Yousif Abuidris,
  Waqas Amin, and Jay Kumar.
\newblock Blockchain-federated-learning and deep learning models for {COVID-19
  detection using CT imaging}.
\newblock {\em arXiv preprint arXiv:2007.06537}, 2020.

\bibitem{vaid2020federated}
Akhil Vaid, Suraj~K Jaladanki, Jie Xu, Shelly Teng, Arvind Kumar, Samuel Lee,
  Sulaiman Somani, Ishan Paranjpe, Jessica~K De~Freitas, Tingyi Wanyan, et~al.
\newblock Federated learning of electronic health records improves mortality
  prediction in patients hospitalized with covid-19.
\newblock {\em medRxiv}, 2020.

\bibitem{li2020federated}
Tian Li, Anit~Kumar Sahu, Ameet Talwalkar, and Virginia Smith.
\newblock Federated learning: Challenges, methods, and future directions.
\newblock {\em IEEE Signal Processing Magazine}, 37(3):50--60, 2020.

\bibitem{li2019differentially}
Jeffrey Li, Mikhail Khodak, Sebastian Caldas, and Ameet Talwalkar.
\newblock Differentially private meta-learning.
\newblock {\em arXiv preprint arXiv:1909.05830}, 2019.

\bibitem{melis2019exploiting}
Luca Melis, Congzheng Song, Emiliano De~Cristofaro, and Vitaly Shmatikov.
\newblock Exploiting unintended feature leakage in collaborative learning.
\newblock In {\em 2019 IEEE Symposium on Security and Privacy (SP)}, pages
  691--706. IEEE, 2019.

\bibitem{qayyum2020secure}
Adnan Qayyum, Junaid Qadir, Muhammad Bilal, and Ala Al-Fuqaha.
\newblock Secure and robust machine learning for healthcare: A survey.
\newblock {\em IEEE Reviews in Biomedical Engineering}, 2020.

\bibitem{qayyum2020securing}
Adnan Qayyum, Muhammad Usama, Junaid Qadir, and Ala Al-Fuqaha.
\newblock Securing connected \& autonomous vehicles: Challenges posed by
  adversarial machine learning and the way forward.
\newblock {\em IEEE Communications Surveys \& Tutorials}, 22(2):998--1026,
  2020.

\bibitem{fang2019local}
Minghong Fang, Xiaoyu Cao, Jinyuan Jia, and Neil~Zhenqiang Gong.
\newblock Local model poisoning attacks to byzantine-robust federated learning.
\newblock {\em arXiv preprint arXiv:1911.11815}, 2019.

\bibitem{10.3389/fdata.2020.587139}
Adnan Qayyum, Aneeqa Ijaz, Muhammad Usama, Waleed Iqbal, Junaid Qadir, Yehia
  Elkhatib, and Ala Al-Fuqaha.
\newblock Securing machine learning in the cloud: A systematic review of cloud
  machine learning security.
\newblock {\em Frontiers in Big Data}, 3:43, 2020.

\bibitem{zhu2020toward}
Guangxu Zhu, Dongzhu Liu, Yuqing Du, Changsheng You, Jun Zhang, and Kaibin
  Huang.
\newblock Toward an intelligent edge: wireless communication meets machine
  learning.
\newblock {\em IEEE Communications Magazine}, 58(1):19--25, 2020.

\bibitem{gobieski2019intelligence}
Graham Gobieski, Brandon Lucia, and Nathan Beckmann.
\newblock Intelligence beyond the edge: Inference on intermittent embedded
  systems.
\newblock In {\em Proceedings of the Twenty-Fourth International Conference on
  Architectural Support for Programming Languages and Operating Systems}, pages
  199--213. ACM, 2019.

\bibitem{nagpaldeep}
Chirag Nagpal.
\newblock Deep multimodal fusion of health records and notes for multitask
  clinical event prediction.
\newblock {\em 31st Conference on Neural Information Processing Systems (NIPS
  2017), Long Beach, CA, USA.}, 2017.

\bibitem{zhou2018robust}
Zhenyu Zhou, Haijun Liao, Bo~Gu, Kazi Mohammed~Saidul Huq, Shahid Mumtaz, and
  Jonathan Rodriguez.
\newblock Robust mobile crowd sensing: When deep learning meets edge computing.
\newblock {\em IEEE Network}, 32(4):54--60, 2018.

\bibitem{qayyum2020single}
Adnan Qayyum, Waqas Sultani, Fahad Shamshad, Junaid Qadir, and Rashid Tufail.
\newblock Single-shot retinal image enhancement using deep image priors.
\newblock In {\em International Conference on Medical Image Computing and
  Computer-Assisted Intervention}, pages 636--646. Springer, 2020.

\bibitem{booz2017decentralized}
David~A Booz, Jonathan~D Dye, Michael~J Dye, and Egan~F Ford.
\newblock Decentralized autonomous edge compute coordinated by smart contract
  on a blockchain, September~28 2017.
\newblock US Patent App. 15/082,559.

\bibitem{psaras2018decentralised}
Ioannis Psaras.
\newblock Decentralised edge-computing and {IoT} through distributed trust.
\newblock In {\em Proceedings of the 16th Annual International Conference on
  Mobile Systems, Applications, and Services}, pages 505--507. ACM, 2018.

\bibitem{zhao2020privacy}
Yang Zhao, Jun Zhao, Linshan Jiang, Rui Tan, Dusit Niyato, Zengxiang Li,
  Lingjuan Lyu, and Yingbo Liu.
\newblock Privacy-preserving blockchain-based federated learning for iot
  devices.
\newblock {\em IEEE Internet of Things Journal}, 2020.

\bibitem{howard2017mobilenets}
Andrew~G Howard, Menglong Zhu, Bo~Chen, Dmitry Kalenichenko, Weijun Wang,
  Tobias Weyand, Marco Andreetto, and Hartwig Adam.
\newblock Mobilenets: Efficient convolutional neural networks for mobile vision
  applications.
\newblock {\em arXiv preprint arXiv:1704.04861}, 2017.

\bibitem{iandola2016squeezenet}
Forrest~N Iandola, Song Han, Matthew~W Moskewicz, Khalid Ashraf, William~J
  Dally, and Kurt Keutzer.
\newblock {SqueezeNet: AlexNet-level accuracy with 50x fewer parameters and<
  0.5 MB model size}.
\newblock {\em arXiv preprint arXiv:1602.07360}, 2016.

\bibitem{han2015learning}
Song Han, Jeff Pool, John Tran, and William Dally.
\newblock Learning both weights and connections for efficient neural network.
\newblock In {\em Advances in neural information processing systems}, pages
  1135--1143, 2015.

\bibitem{wen2016learning}
Wei Wen, Chunpeng Wu, Yandan Wang, Yiran Chen, and Hai Li.
\newblock Learning structured sparsity in deep neural networks.
\newblock In {\em Advances in neural information processing systems}, pages
  2074--2082, 2016.

\bibitem{yim2017gift}
Junho Yim, Donggyu Joo, Jihoon Bae, and Junmo Kim.
\newblock A gift from knowledge distillation: Fast optimization, network
  minimization and transfer learning.
\newblock In {\em Proceedings of the IEEE Conference on Computer Vision and
  Pattern Recognition}, pages 4133--4141, 2017.

\bibitem{sattler2019clustered}
Felix Sattler, Klaus-Robert M{\"u}ller, and Wojciech Samek.
\newblock Clustered federated learning: Model-agnostic distributed multi-task
  optimization under privacy constraints.
\newblock {\em arXiv preprint arXiv:1910.01991}, 2019.

\bibitem{horry2020covid}
Michael~J Horry, Subrata Chakraborty, Manoranjan Paul, Anwaar Ulhaq, Biswajeet
  Pradhan, Manas Saha, and Nagesh Shukla.
\newblock {COVID-19} detection through transfer learning using multimodal
  imaging data.
\newblock {\em IEEE Access}, 8:149808--149824, 2020.

\bibitem{lin2017focal}
Tsung-Yi Lin, Priya Goyal, Ross Girshick, Kaiming He, and Piotr Doll{\'a}r.
\newblock Focal loss for dense object detection.
\newblock In {\em Proceedings of the IEEE international conference on computer
  vision}, pages 2980--2988, 2017.

\bibitem{finlayson2019adversarial}
Samuel~G Finlayson, John~D Bowers, Joichi Ito, Jonathan~L Zittrain, Andrew~L
  Beam, and Isaac~S Kohane.
\newblock Adversarial attacks on medical machine learning.
\newblock {\em Science}, 363(6433):1287--1289, 2019.

\bibitem{mirsky2019ct}
Yisroel Mirsky, Tom Mahler, Ilan Shelef, and Yuval Elovici.
\newblock Ct-gan: Malicious tampering of 3d medical imagery using deep
  learning.
\newblock In {\em 28th USENIX Security Symposium USENIX Security 19)}, pages
  461--478, 2019.

\bibitem{rahman2020adversarial}
Abdur Rahman, M~Shamim Hossain, Nabil~A Alrajeh, and Fawaz Alsolami.
\newblock Adversarial examples--security threats to {COVID-19 deep learning
  systems in medical IoT} devices.
\newblock {\em IEEE Internet of Things Journal}, 2020.


\end{thebibliography}

\end{document}